\documentclass[a4paper]{article}
\usepackage{amsmath}
\usepackage{color}
\usepackage{INTERSPEECH2018}

\title{Measuring Conversational Productivity in Child Forensic Interviews}
\name{Victor Ardulov$^1$, Manoj Kumar$^1$, Shanna Williams$^2$, Thomas Lyon$^2$, Shrikanth Narayanan$^1$}
\address{
  $^1$ Signal Analysis and Interpretation Lab\\
  $^2$ Child Interviewing Laboratory\\
  University of Southern California}
\email{ardulov@usc.edu}
\begin{document}
\maketitle
\begin{abstract}
Child Forensic Interviewing (FI) presents a challenge for effective information retrieval and decision making. The high stakes associated with the process demand that expert legal interviewers are able to effectively establish a channel of communication and elicit substantive knowledge from the child-client while minimizing potential for experiencing trauma. 
As a first step toward computationally modeling and producing quality spoken interviewing strategies and a generalized understanding of interview dynamics, we propose a novel methodology to computationally model effectiveness criteria, by applying summarization and topic modeling techniques to objectively measure and rank the responsiveness and conversational productivity of a child during FI.
We score information retrieval by constructing an agenda to represent general topics of interest and measuring alignment with a given response and leveraging lexical entrainment for responsiveness.
For comparison, we present our methods along with traditional metrics of evaluation and discuss the use of prior information for generating situational awareness.

\end{abstract}
\section{Introduction}
A Forensic Interview (FI) with a child involves a legal expert navigating a semi-structured interaction with an objective to elicit substantive and detailed information regarding abuse or neglect that the child might have witnessed or been a victim of. Given the potential criminal underpinning, risks, and consequences associated with these spoken interviews, it is critical they are conducted to evoke genuine information from a possible sufferer. We explore the role of speech and language processing to provide supporting analytical tools in this domain. 

Since FI is conducted with a potential victim of criminal maltreatment, it runs the risk of emotionally impacting the child (e.g. potentially having a victim re-experience trauma). Establishing a sense of trust and comfort with the children strongly impacts the outcome of the interviews \cite{sternberg1997effects,lamb2009use}. FI with children is further challenged as the child's age and cognitive development and state may impact their ability to provide a complete and accurate response. For instance, \cite{Lyon2007} demonstrated that sexually-abused children were more likely to deny the abuse during their first line of questioning. This imposes a tremendous importance on developing strategies for rapport building with the child which can adequately capture the multi-faceted complexity of the objective and constraints.

In order to effectively research and develop best practices for FI, we find it necessary to have an objective method to assess quality of interactions. A number of contemporary studies \cite{ahern2015prosecutors,price2016rapport,talwar2018does} have evaluated interaction using a measure of ``productivity" scoring of child responses. The dominant metrics used are utterance word count along with a variety of human-coded measures such as ``richness" \cite{Lamb1996} indicating the number of informative details provided \cite{Ahern2015}. In this paper, we address the limitations of these methods and present a novel approach to identify momentary instances of verbal productivity and evaluate for each utterance. 

This work both builds upon existing NLP methods and presents new metrics to computationally model topics presented in goal-oriented dialogue, which extend to a mathematical formulation and framework for extracting the conversational ``agenda" to represent information that the interviewer wishes to address. We present a number of scoring methods which leverage both a child's responsiveness to an immediate prompt as well as the alignment of utterances with the agenda. To the best of our knowledge, this is the first approach to scoring verbal recall productivity using computational metrics, and believe that it has the potential to strongly impact the field of FI, giving researchers an objective measure which they can optimize, score, and inform their decision making.

\section{Background}
FI of children is an important and active area of research in psychology and law. The situational context is high stakes as children are frequently the victims of crimes perpetrated by their caretakers or legal guardians \cite{radford2011child}, creating a conflict for both interviewers and the child. Furthermore, children are often the only witnesses to the abuse of others \cite{Lamb2003} making their recall and accounts provided during interviews crucial to the legal process. Speech and language analysis performed in \cite{Ahern2015} and \cite{Klemfuss2015} provides details on the effectiveness across linguistic features for questions and show that contemporary interviewing research is conducted using word count and human coding of new relevant information	. 

\begin{figure}[h]
\centering
\includegraphics[width=.80\linewidth]{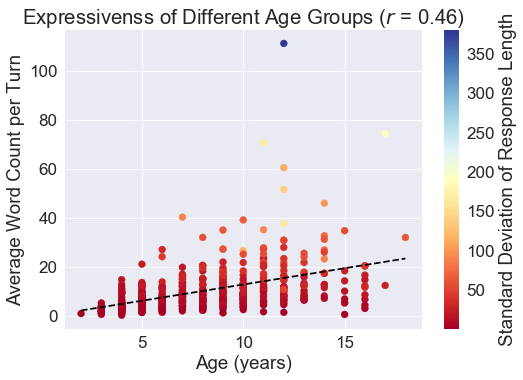}
\caption{Expressiveness (in terms of average turn-level word count) of children across age groups collected over 527 Forensic Interview Transcripts}
\label{fig:child_expressiveness}
\end{figure}

Word count is generally recorded as a metric of responsiveness during rapport building.  Aside from establishing a comfortable setting, rapport building phases are important to practicing narrative techniques in children \cite{Anderson2014}. However, Figure \ref{fig:child_expressiveness} shows that the expressiveness in terms of per-turn average word count and its variability per group grows with a steady relation to their age. Generally though, we see that most children will have relatively low variance in the number of words they express during a session. Thus, word count is more likely to be reflective of an individual's language usage and ability, rather than a indicator of expressiveness or narrative structuring. By introducing a computational metric to capture responsiveness as a measure of topic expression we hope to facilitate the evaluation with higher fidelity. Furthermore, the ability to automate the interview evaluation alleviates some of the need for human involvement in data creation and labeling, making the process more scalable, faster, and less subjective.

The importance of social cues and rapport during an interview make it similar in nature to psychotherapeutic (PT) dialogues such as motivational interviewing, where the interviewer (counselor) is guiding the subject (client) towards a goal through appropriate dialogue actions\cite{Lord2015}. Both FI and PT require a foundation of trust, which is built over the interactions and do not always align with the direct objectives of the dialogue. In psychotherapy trust is used to promote behavioral change in the client, typically over a series of sessions over a long period of time. However, the legal setting surrounding FI limits engagement opportunities that interviewers have, and is more focused in maximizing information elicitation from the subject without coercion.

Previous work on dialogue-summarization shows success in tracking and modeling topics discussed in dyadic conversations. One source of inspiration for our work is the topic modeling performed in \textit{DiaSumm} \cite{Zechner2000}, which uses \textit{tf-idf} and TextTiling \cite{Hearst1997} to find topics and their boundaries in the CALLHOME dataset. However, since we tasked with scoring the relevance of a response, our approach only uses the interviewers side of the dialogue to extract the topics.

The work presented in \cite{Noh2013} takes advantage of knowledge-graphs (e.g. Wikipedia) to improve dialogue summary and topic identification robustness. However within the specific application of Child FI, the vocabulary used by the client will be strongly influenced by the child's language development. As a result, semantic representations from these large knowledge-graphs do not necessarily contain information that can improve the topic-modeling of the dialogues.

Finally, both Question-Answering Systems (QASs) and Machine Dialogue Systems (MDSs) provide frameworks for evaluating machine-generated responses. Many different flavors exist for QASs, both Natural-Language-Understanding based and structured-query, but both are assessed in whether or not the response was correct in reference to a ground truth \cite{Bouziane2015}. MDSs are generally evaluated for ``human-like'' linguistic features using word-perplexity as a first pass, and then relying on human analysis of the generated response \cite{Serban2016, Li2017}. When evaluating goal-oriented dialogue systems such as \cite{Lewis2017}, the response is scored by the final objective measure of the system's ability to outperform a human in a negotiation. Ultimately, both of these domains present relevant methodologies towards how a dialogue may be scored, but neither present a specific criteria that may be applied to the context of Child FI. 

In juxtaposition, this paper presents a scoring metric which evaluates dialogue interactions without a known ground truth which we hope to elicit, nor a known objective metric that is being maximized through the interaction. The methodologies presented in Sections \ref{section:agenda} and \ref{section:responsiveness} address the limitations of human-coding scalability and word count applicability. Our results demonstrate that despite these challenges a computationally motivated metric generates robust and informative values indicative of interview effectiveness.

\section{Methodology}
Let us introduce some notation to begin. An interview $\Psi = \{(q_0, r_0), \hdots, (q_m, r_m)\}$ is denoted as a sequence of questions ($q_t$) and response phrases ($r_t$). An interviewer has a vocabulary $\mathcal{V}_{\Psi}$ which is a collection of unique n-grams for an interviewer in $\Psi$. We chose arbitrary n = 3 and exclude stop words as they by definition do not carry meaning.

\subsection{Agenda}
\label{section:agenda}
We define an agenda, for an interview session $\Psi$ in the following way
$$ \mathcal{A}_{\Psi} = \phi(\mathcal{V}_{\Psi}, Q_\Psi) = \langle tf(v_j, Q_{\Psi}) \rangle_{\forall v_j \in \mathcal{V}_{\Psi}}$$

Where $\phi$ is a function to construct a vector of term-frequency, $tf(v, Q_{\Psi})$,  of word $v$ in $Q_{\Psi} = \bigcup_{\forall t} q_t$.

We construct a vector to represent a response phrase $r_t$ within the interviewers vocabulary $\mathcal{V}_{\Psi}$:
$$ \vec{\textbf{r}}_t = \phi(\mathcal{V}_{\Psi}, r_)$$
From this we present our first productivity score ($g$) of a given response as:
$$ g(r_t) = \textbf{r}_t \cdot \mathcal{A}_{\Psi} $$
We will refer to this as the Agenda productivity, and justify that $\mathcal{A}_\Psi$ is a representation of the topics, that are most relevant to the dialogue. By taking the dot product we compare how much of the agenda is expressed in the response as time $r_t$. This does not discredit the word count, however we apply less weight to utterances that are not relevant to the topics we believe require information.

\subsection{Responsiveness}
\label{section:responsiveness}
Observe that the agenda is constructed over the entire session. However, during a dialogue the agenda becomes revealed over time, implying that early interactions may be given low scores inherently because not enough of the agenda has been revealed to the subject in order for them to respond meaningfully. Simultaneously, there is a desire to assign productivity to the response elicited if the child responded directly to a prompt from the interviewer. We will use this as inspiration to construct a term that will give benefit to responsiveness (immediate address of prompted agenda topic). Let us then construct $a_t$ which is the \textit{rolling agenda}, a measure of the agenda that has been revealed:
$$a_t = \phi(\mathcal{V}_{\Psi}, q_t) + \gamma a_{t-1} =\sum_{i = 0}^{t} \gamma^{i} \phi(\mathcal{V}_{\Psi}, q_{t-i})$$ 
This allows us to treat $a_t$ as a representation of recently evoked words using $\gamma \in [0, 1]$ to represent a discount on words or topics that are brought up in the past. Alignment or use of these words can be observed as a lexical entrainment or responsiveness to the question, which are indicators of trust between the interviewer and the client. Entrainment refers to instances when conversational partners begin to use the same vocabulary and match each other's speaking patterns \cite{Brennan1996}. We define an instance of entrainment or responsiveness as the reuse of vocabulary from a previous point in time as:
$$
\rho(r_t) = \textbf{r}_t \cdot a_t
$$
Then to capture both of the desired scores we construct the following productivity measure:
$$\pi^*(r_t) = \beta \frac{\rho(r_t)}{||a_t||} + (1 - \beta) \frac{g(r_t)}{||\mathcal{A}||} $$
where $\beta \in [0, 1]$ allows us to flexibly
leverage the complementary importance of the skills being captured by each sub-metric.

\begin{figure}[h]
\centering
\includegraphics[width=.8\linewidth]{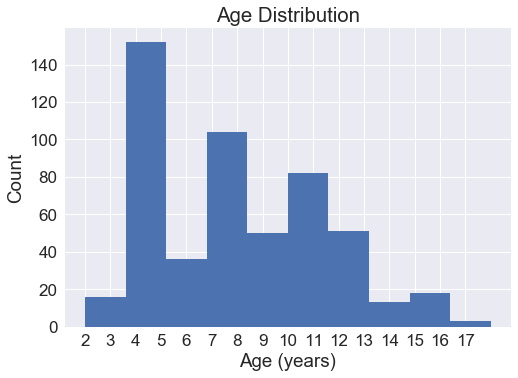}
\caption{Age distribution across 527 sessions}
\label{fig:age_distro}
\end{figure}

\section{Results and Discussion}
Below, we apply the considered productivity scoring functions across a set of 527 child forensic interviews. Each data-point represents a single interview between an expert-legal interviewer and a unique child (i.e. each interview is with a different child). The children interviewed are aged between 2 and 17 years old with the exact distribution shown in Figure \ref{fig:age_distro}. The data are collected as part of a study for the development of interviewing practices and contain examples of various interviewing strategies. To validate our methods we present samples randomly selected from the dataset, highlighting ability of our method to extract important concepts and the differences observed across the use of 4 different scoring metrics: word count, agenda ($g$) scoring, responsiveness ($\rho$) scoring, and a combined responsiveness/agenda ($\pi^*$) scoring.
\subsection{Agenda and Responsiveness}
\begin{table}[h]
\centering
\caption{Top 10 weighted words from 5 agendas}
\label{table:agenda_example}
\begin{tabular}{c c c c c}
\hline
$\Psi_1$ & $\Psi_2$ & $\Psi_3$ & $\Psi_4$ & $\Psi_5$\\
\hline
bathroom & mister & cousin & gonna & really\\
gonna & pinched & al & mommy & understand\\
outside & kids & cousin al & aunt & gonna\\
touched & thank & touched & touched & sometimes\\
garage & gonna & private & touching & times\\
uncle & mom & talking & michael & important\\
sitting & wrong & doors & clothes & clothes\\
clothes & kid & outside & peepee & kids\\
fix & id & gonna & pants & mom\\
mom & really & legs & grandma & touched\\
\hline
\end{tabular}
\end{table}

Table \ref{table:agenda_example} shows examples of agendas that have been constructed using our methods with simple stop-word filtering.
The extraction clearly identifies important concepts that are relevant to potential episodes of abuse that the subject experiences or witnesses. Particularly we take note that these words were extracted purely from the lexicon of the interviewer, and are not directly modeling the response of the child. This is justified by the assumption that in an interview setting, a concept that is repeated frequently in the questions is indicative of a broader interest in that topic.

\begin{table*}[h]
\caption{Comparison of productivity produced by different methods and their relative rank (1 being highest rated rank). Demonstrates the ability to score substantive information from utterances. For Responsiveness+Agenda hyper-parameters $\gamma$ and $\beta$ are chosen to be 0.5}
\label{table:utter_example}
\centering
\begin{tabular}{l c c c c}
\hline
\hfill &\textbf{Word Count}&\textbf{Agenda} & \textbf{Responsive}&\textbf{Responsiveness/Agenda}\\
\textbf{Utterance Excerpt} &\hfill & ($g$) & ($\rho$) & ($\pi^*$)\\
\hline
I like to play with my friends. & & & &\\
I have a new student ... & 35 (1) & 0.00 (128) & 0.00 (129) & 0.00 (128)\\
\hline
Outside in the bathroom... & & & &\\ 
(child reveals details) & 9 (26) & 27.00 (1) & 0.16 (3) & 0.23 (2) \\
\hline
Over my skirt clothes, & & & \\
but outside he got me under my clothes. & 12 (17) & 12.00 (3) & 0.37 (1) & 0.40 (1) \\
\hline
\end{tabular}
\end{table*}

A closer exploration of utterance level interactions and their scoring can be found in Table \ref{table:utter_example}, which exhibits the differences and similarities between scoring metrics. Comparing the rating of each method against the others' top-rated utterances informs us as to what features each criterion is sensitive to.

When using the word count, a response to the interviewer asking the child to talk about their experience in school is rated highly, despite not being substantive to overall interview. While the response does generally contain a lot of words (and thus information) and can be seen as an demonstration of trust, it does not directly indicate any clues relevant to the suspected abuse the child might have experienced. In contrast with word count, our scoring metrics assign higher score to utterances that contain information relevant to more targeted questions. 

This is not to say that asking about the child's schooling is unproductive to the interview. Rather, the information explored in that specific utterance is not indicative of the overall interview session's effectiveness. The use of the agenda allows us to assign higher productivity to utterances in which the child reveals a strong cooperation with the goals of the interviewer. When taken in conjunction with the possibility of value-iteration methods, we allow for a decision theoretic model to assign value to rapport building dialogue actions which will optimize for the desired objective of building responsive communication and eliciting information.

\subsection{Signal Sparsity}
\begin{figure}[h]
\centering
\includegraphics[width=.45\linewidth]{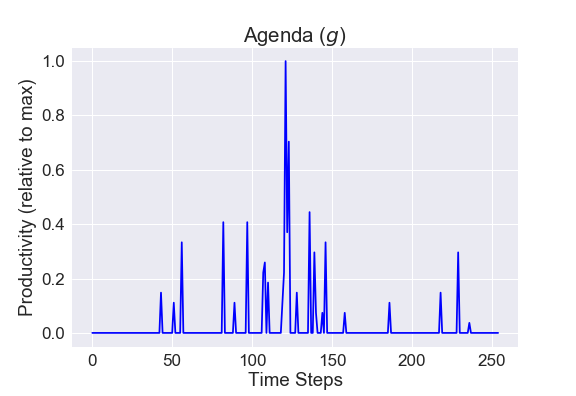}
\includegraphics[width=.45\linewidth]{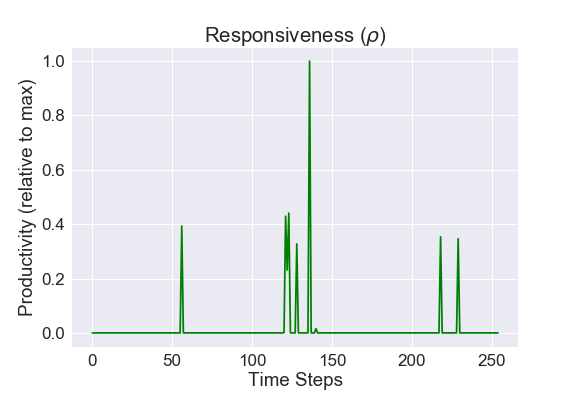}
\includegraphics[width=.45\linewidth]{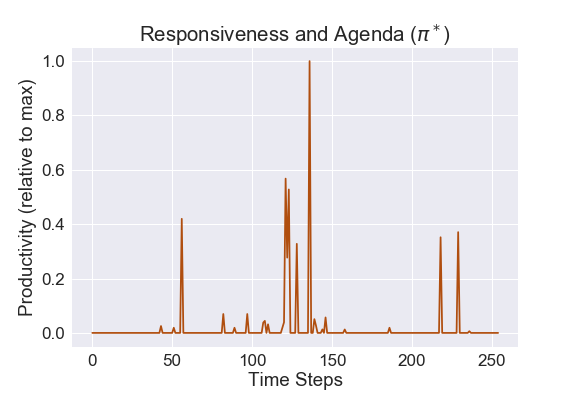}
\includegraphics[width=.45\linewidth]{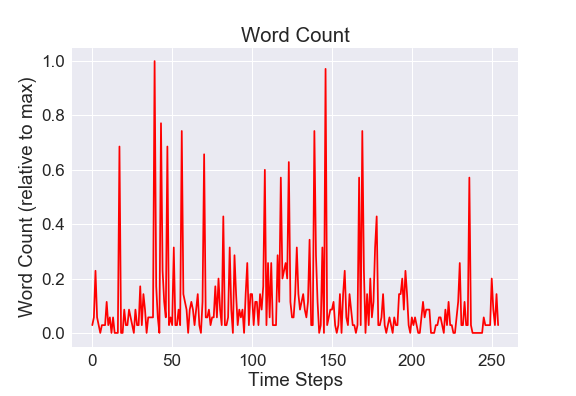}
\caption{Various proposed productivity scores over the course of a single session, agenda(\textit{top, left}), responsiveness (\textit{top, right}), combined (\textit{bottom, left}) and word count (\textit{bottom, right}). Each score is normalized with respect to its maximum.}
\label{fig:prod_over_time}
\end{figure}
By comparing the differences in signal sparsity shown in Figure \ref{fig:prod_over_time}, it can be seen that the word count metric, while being highly abundant, also demonstrates a highly volatile nature suggesting many hidden underlying influences. The signal produced by the productivity metric, while sparser, can be interpreted as a information-relevance filter convolved against the lexical information provided by the child. This implies that sparsity will not hinder the study of productivity, instead allows us to isolate the effects of differing strategies specifically to the information retrieval criteria.
\begin{table}
\centering
\caption{Pearson correlation (\textit{r}) indicating strength of relationship between different productivity metrics and the age of the child. All scores were significant at $p < 0.001$.}
\begin{tabular}{c c}
\hline
\textbf{Metric}& \textit{r} \\
\hline
Word Count & 0.46\\
Agenda $g$ &0.26 \\
Responsive $\rho$ & 0.24\\
Responsive/Agenda $\pi^{*}$ & 0.25\\
\hline
\end{tabular}
\label{table:pearson_scores}
\end{table}

\subsection{Age-Related Analysis}

\begin{figure}[h]
\centering
\includegraphics[width=.45\linewidth]{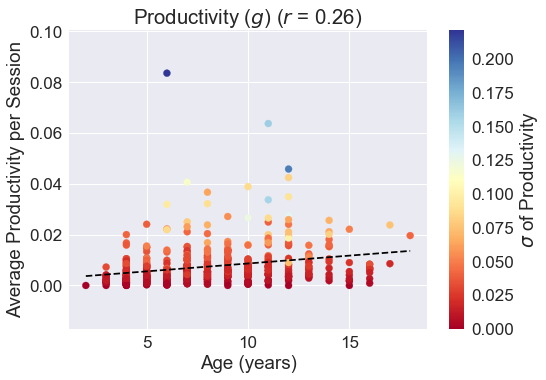}
\hspace{0.04\linewidth}
\includegraphics[width=.45\linewidth]{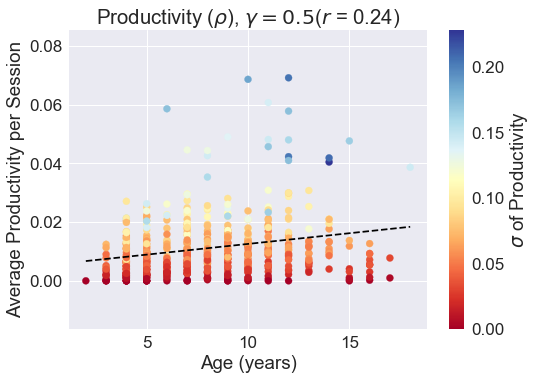}
\caption{Distribution of scores and variances by age using only the Agenda criterion}
\label{fig:g_rho_prod_distros}
\end{figure}

\begin{figure}
\centering
\includegraphics[width=.80\linewidth]{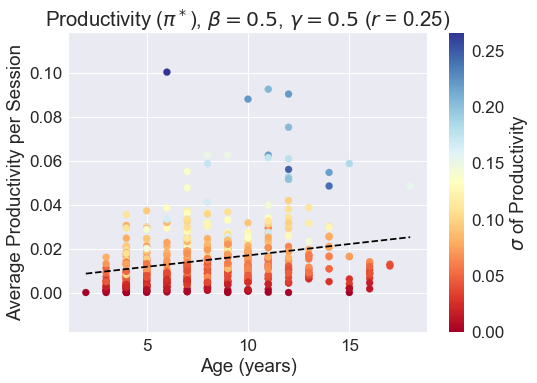}
\caption{Distribution of scores and variances by age using Responsiveness scoring methods.}
\label{fig:pi_prod_distros}
\end{figure}

For comparison with Figure \ref{fig:child_expressiveness}, Figures \ref{fig:g_rho_prod_distros} and \ref{fig:pi_prod_distros} show us the resulting dependency between productivity scores and child age. Moreover, Table \ref{table:pearson_scores} reveals the difference in Pearson scores to be statistically significant, implying a weaker correlation between age and productivity, further suggesting that the metrics are also more robust to a specific child's language ability and development. From this we can extrapolate that these metrics can be used to evaluate the productivity adaptively to each individual and capture a more general set of interaction characteristics. 

\section{Conclusions}
Our experiments and evaluations demonstrate the usefulness of using topic modeling for productivity assessment. We further suggest the value captured by measuring responsiveness is equivalent to a time-dilated matching of vocabulary usage corresponding to lexical entrainment of local interactions, a known indicator of trust in a dyadic interaction \cite{Scissors2009}. A number of promising examples for these methods have been shown to significantly improve productivity measurements of Child FI while also automating the process. Furthermore, the results demonstrate that this can be achieved using a rather simple trigram model.

By introducing hyper-parameters $\beta$ and $\gamma$ to the proposed productivity measure we allow the system to maintain flexibility for different criteria,  without sacrificing consistency of scoring across specific examples.  For the first time, FI researchers will be able to rigorously assign a value to a response within the context of a conversation. Aside from simply reevaluating prior work, these techniques allow for interviewers to construct ``gold standard" agendas prior to the interview. Using these prepared agendas will allow interviewers to use situational hypotheses and prior knowledge to improve and inform the specific policy or strategy they employ.

More broadly, these techniques can be used to model similar goal-oriented dialogues such as negotiation and question answering in an effort to make information discovery a larger part of the dialogue process.

In conclusion, our results suggest that the computational models of these interviews are better informed by the proposed measure of productivity. This creates a clear criterion that can be evaluated and optimized, providing real-time updates and adaptive scoring metrics.
\section{Acknowledgements}
We would like to thank our collaborators at the Child Interviewing Lab at the Gould School of Law for their collection and sharing of the data presented in this paper.

\bibliographystyle{IEEEtran}
\bibliography{mybib.bib}
\end{document}